\documentclass[a4paper, 10pt, conference]{ieeeconf}      
\usepackage{FG2024}

\FGfinalcopy 

\IEEEoverridecommandlockouts                              
\usepackage{amsmath} 
\usepackage{amssymb}  
\usepackage{adjustbox}
\usepackage{mathtools}
\usepackage{times}
\usepackage{epsfig}
\usepackage{graphicx}
\usepackage{booktabs}
\usepackage{graphicx}
\usepackage{subcaption}

\makeatletter
\let\NAT@parse\undefined
\makeatother
\usepackage[numbers]{natbib}
\def\FGPaperID{40} 

\title{\LARGE \bf Efficient Verification-Based Face Identification}

\author{\parbox{16cm}{\centering
    {\large Amit Rozner$^{\ast1}$ Barak Battash$^{\ast1}$ Ofir Lindenbaum$^1$ and Lior Wolf$^2$}\\
    {\normalsize
    $^1$ Faculty of Engineering, Bar Ilan University\\
    $^2$ School of Computer Science, Tel Aviv University\\
    $^\ast$ These authors contributed equally.\\}}
}
\usepackage{fancyhdr}
\begin{document}

\ifFGfinal
\thispagestyle{empty}
\pagestyle{empty}
\else
\author{Anonymous FG2024 submission\\ Paper ID \FGPaperID \\}
\pagestyle{plain}
\fi
\maketitle

\thispagestyle{fancy} 

\begin{abstract}
We study the problem of performing face verification with an efficient neural model $f$. The efficiency of $f$ stems from simplifying the face verification problem from an embedding nearest neighbor search into a binary problem; each user has its own neural network $f$. To allow information sharing between different individuals in the training set, we do not train $f$ directly but instead generate the model weights using a hypernetwork $h$. This leads to the generation of a compact personalized model for face identification that can be deployed on edge devices. Key to the method's success is a novel way of generating hard negatives and carefully scheduling the training objectives. Our model leads to a substantially small $f$ requiring only 23k parameters and 5M floating point operations (FLOPS). We use six face verification datasets to demonstrate that our method is on par or better than state-of-the-art models, with a significantly reduced number of parameters and computational burden. Furthermore, we perform an extensive ablation study to demonstrate the importance of each element in our method.

\end{abstract}
\section{Introduction}
Face recognition is often tackled as a multiclass classification task in which a different class represents each identity. However, this paradigm leads to fixed identities based on those observed during training. A more general approach is to perform matching using the learned representation~\cite{deepface}. This basic form of transfer learning has proven scalable and relatively efficient~\cite{webscale}. 

Sometimes, however, the resources needed to compute the representation are infeasible, e.g., for edge devices with limited resources. Common restrictions involve limited memory footprint and energy consumption. Various edge devices require different types of optimizations. For example, mobile phones, laptops, and tablets require energy consumption optimization to achieve enhanced battery life, which is a top priority for such devices \cite{ajani2021overview}. Additional challenges are often faced in IP cameras, smart doors, and other internet-of-things (IoT) devices with low compute and memory resources \cite{rastegari2020enabling,zemlyanikin2019512kib}.

The prevailing approach for addressing the face recognition challenge involves employing a large overparameterized deep neural network (DNN). Such a network is proficient at converting a facial image into a vector within an embedding space. Each identity has at least one face image, represented using an embedding vector in the database. During the inference phase, each facial image must undergo processing through an extensive DNN to generate an embedding vector, facilitating subsequent comparisons with vectors stored in the database. Consequently, the computational expense of the inference operation is influenced directly by the complexity of the DNN. Prior research efforts have proposed using more lightweight DNN architectures to mitigate this computational burden. In this research project, we are interested in the face verification setting, in which we want to learn a model specific to one or a few identities.


In such cases, it makes sense to consider a verification-based paradigm, in which one trains much smaller networks, each designed to identify a specific individual, thus simplifying the face identification problem. Suppose the number of individuals in the gallery is not large. In this case, this ensemble of binary classifiers may be more efficient than the representation-based approach, especially on edge devices, such as mobile phones, where only a single user is enrolled in the system. 

Training multiple binary classifiers, however, has an inherent limitation in that the benefit of knowledge sharing is mainly lost. Face identification requires invariance to pose, illumination, expression, background, and other factors. When training a deep multiclass classifier, this invariance is manifested in the various layers, and learning to be pose-invariant on the images of one person, for example, can be beneficial for identifying others. 
Further, the binary classifier needs to be trained for each enrolled user. This results in a prolonged ``sign-up'' process with additional computational costs. Fine-tuning deep neural networks using a single or a few user images is challenging, even if the backbone is pre-trained.

Some knowledge sharing is possible through shared layers in the ensemble of binary classifiers. However, knowledge sharing occurs at all levels (both low-level vision and high-level). Moreover, due to the application demands, the binary networks need to be of low capacity, and sharing some of the layers would reduce the ability to adapt to the new identities.

In this work, we suggest an alternative solution in which a hypernetwork~\cite{ha2016hypernetworks} $h$ is trained to generate, given a single image $x^i_{enroll}$ of a person $i$, the weights $\theta^i$ of a verification network $f(\theta^i)$ that identifies the specific individual $i$. Our model is trained using weighted binary cross-entropy loss, and we introduce a new method for effective hard negative mining. The new scheme, termed $K$-means centered sampling (KCS), helps our model learn a more discriminative function by constructing batches containing similar identities. We benchmark our method using six publicly available datasets and demonstrate that it can dramatically reduce the number of parameters while maintaining highly competitive results.

In the following sections, we detail the background of our work \ref{sec:background} as well as its related research \ref{sec:related_work}. Then, we provide a detailed description of our method in section \ref{method}. Following this, we present the experiments we conducted in section \ref{sec:experiments}, which is followed by a discussion \ref{sec:discussion} and conclusions \ref{sec:conclusions} sections. 

\section{Background}
\label{sec:background}
Facial recognition involves recognizing a person enrolled in a database by their facial image. The two main tasks in facial recognition are face verification (one-to-one matching 1:1) and face identification (one-to-many matching 1:N). Face verification involves comparing two facial images from a database and another non-database image (query image). Then, the system decides whether the person in the query image is of the same identity as in the database image. Face identification involves authenticating a person's identity, considering multiple identities in the database. Since face verification involves a 1-to-1 matching, it is often a more straightforward task, which requires less computational effort and provides higher accuracy than identification.

Face identification is used for surveillance \cite{cheng2018surveillance}, access control \cite{nag2018iot,lee2020face}, law enforcement \cite{lynch2020face}, etc. It often involves a large amount of identities in the database. This may require the model to run on designated hardware with high power and computing capabilities. For some applications, edge devices are power consumption bounded by battery and/or heat limitations. Thus, using huge and power-demanding models is less common in real-world applications. This issue can sometimes be mitigated by offloading the DNN computations to the cloud, which produces more significant latency and additional costs. Face verification is commonly used in granting access to personal mobile devices, tablets, or computers, payment verification, and more. Despite the significant difference between the two tasks, the standard way of solving them is typically based on the same framework.

Deep learning is the main driver behind modern face recognition systems. Many systems are based on a deep neural network $h_{bb}$, which projects an input face image $x_{enroll}$ to an embedding vector $e_{enroll} = h_{bb}(x_{enroll})$, where $e_{enroll}\in R^{d}$ and $d$ is the embedding dimension.
Face recognition consists of two phases: enrollment and authentication. In the enrollment stage, a face image of a person $i$ is fed to $h_{bb}$, generating a unique feature vector $e^i_{enroll}$ based on their unique facial characteristics. $e^i_{enroll}$ is then stored in a database $D\in R^{N\times d}$, where $N$ is the number of faces stored in the database. 

The difference between identification and verification tasks can be formally defined by $N$, where $N>1$, and $N=1$ are used in identification and verification, respectively. For authentication, a face image $x_{auth}$ is fed to the feature extractor $e_{auth} = h_{bb}(x_{auth})$, where $e_{auth}$ is the embedding vector of $x_{auth}$. For the verification task, $e_{auth}$ is compared to $e^i_{enroll}$ using a predefined distance metric, such as the L2 distance, cosine similarity, or similar. Suppose the distance between the two embedding vectors is lower than a threshold. In that case, the face is authenticated as belonging to the same identity, i.e., $x_{auth}$ is an image of person $i$. For identification, the authentication is performed in two phases. The initial search is typically performed using a $K$-Nearest-Neighbours \cite{dudani1976distance} algorithm or similar that identifies the closest embedding $e^{nearest}_{enroll}$ and performs verification between $e^{nearest}_{enroll}$ and $e_{auth}$ using the threshold scheme discussed above.

Since DNNs attempt to embed a large set of diverse faces, the required network size is very demanding. We argue that a paradigm shift in this approach may prove beneficial. Instead of generating embeddings for each input face image, developing a more compact DNN that can authenticate the identity of a specific given face would be advantageous. This simplified approach may significantly reduce model size, minimizing the need for cloud computing and significantly decreasing power consumption, thereby positively impacting battery life.

\section{Related Work}
\label{sec:related_work}
\subsection{Deep Embedding Neural Networks for Facial Recognition}
In recent years, DNNs have been widely used for face recognition tasks. \citet{schroff2015facenet} used a Triplet loss to improve performance on challenging facial recognition datasets. Additional progress was made by \citet{liu2016large}, who suggested an improvement to the well-known Softmax loss, termed L-Softmax loss. It encourages intra-class compactness and inter-class separation between the learned embedding features. \citet{liu2017sphereface} introduced the concept of an angular margin loss combined with the standard Softmax loss function. Removing the Softmax and applying a Cosine margin loss to the target logit helped \citet{wang2018cosface} to achieve superior performance over its predecessors. ArcFace \cite{deng2019arcface} used a geodesic distance over a hypersphere, showing improved discriminative power and a stable training process. Using image quality estimation by the embedding norm, \citet{kim2022adaface} claimed that adapting the ArcFace by each sample's quality will lead to significant gains. Their proposed loss, termed AdaFace, achieved state-of-the-art (SOTA) results on challenging low-quality and high-quality datasets.  

\subsection{Face Recognition on Edge Devices}

Edge devices do not have the power and computing abilities required to run SOTA facial recognition models, such as \cite{han2022survey,khan2022transformers}. This has led to the development of efficient neural networks designed to minimize memory and power consumption while maintaining high accuracy. Creating such efficient models requires considerable effort and may involve methods such as pruning \cite{liu2020pruning,liang2021pruning}, quantization \cite{gholami2022survey}, and hardware-aware neural architecture search \cite{benmeziane2021comprehensive}. In recent years, researchers have made significant progress in developing efficient facial recognition models, downscaling them to sizes as small as half a million parameters \cite{martindez2019shufflefacenet}.

Several attempts were made to convert lightweight generic computer vision models for facial recognition tasks to mitigate this issue. An early attempt was made by \citet{chen2018mobilefacenets} using a modified MobileNetV2 \cite{sandler2018mobilenetv2} architecture trained for the face verification task with the ArcFace \cite{deng2019arcface} loss. They achieved real-time yet competitive results using less than 1M parameters. 
Subsequently, \citet{martindez2019shufflefacenet} suggested multiple architectures that resemble ShuffleNetV2 \cite{ma2018shufflenet} with a parameter range from 0.5M to 4.5M. Then, \citet{martinez2021benchmarking} introduced MobileFaceNetV1, which utilized separable convolution to decrease the computational complexity, and ProxylessFaceNAS, which added the inverted residual block to its neural architecture space, thus increasing efficiency. Recently, \citet{hoo2022convfacenext} achieved further parameter reduction by using modified ConvNeXt \cite{liu2022convnet} blocks. While former works used convolutional neural networks (CNNs), \citet{hoo2023lcam} introduced attention modules, achieving further progress. Another approach by \citet{george2023edgeface} combines CNN with Transformer architectures and a low-rank linear layer to reach further SOTA results when training on the WebFace12M dataset \cite{zhu2021webface260m}.

\subsection{HyperNetwork Neural Network Architecture}
HyperNetwork, a term introduced in \cite{ha2016hypernetworks}, is a general name for a neural network model that predicts the weights of another neural network, usually referred to as a target network or a primary network. The target network performs the actual task for which the training was done. Since its inception, it has been a highly successful architecture for many learning tasks, such as uncertainty estimation via Bayesian inference \cite{krueger2017bayesian}, generating weights for DNNs of related tasks, i.e., continual learning \cite{von2019continual}, enabling better compression in resource-limited use cases compared to embedding neural networks \cite{zhao2020meta,nguyen2021fast,galanti2020modularity}, training multiple target DNNs faster \cite{navon2020learning}, and more \cite{yang2022locally,svirsky2023interpretable}. 
We want to use a hypernetwork as a facial recognition classifier, where the target model will be deployed on the edge. 

\section{Method}
\label{method}
Our method consists of two main parts: a single hypernetwork $h$ located on the cloud (while not obligatory), which generates weights $\theta^i$ for a personal neural network $f(\theta^i)$, where $i$ is the user ID. Each user's interaction with hypernetwork $h$ is a one-time occasion only during enrollment. The hypernetwork $h$ comprises two blocks: $h_{gen}$ and $h_{bb}$, in the following manner:
\begin{equation}
    h = h_{gen} \circ h_{bb}.
\end{equation}
The first is a pre-trained face recognition embedding model $h_{bb}$, which is frozen for the entire training process. This block helps us extract face features using a strong pretrained DNN. The second part, $h_{gen}$, is a DNN that receives face information produced by $h_{bb}$ and generates weights $\theta^i$ for a user's personal model $f(\theta^i)$. Next, we rigorously define each component in our system while describing the three phases: enrollment, inference, and training. Although the training phase is the first chronologically, we start from the enrollment and inference phases to better motivate our novel suggested paradigm.

\subsection{Enrollment Phase}
The enrollment phase is the process of inserting an identity $i$ into the database. This enables comparing new unknown images to the enrolled identity and making decisions, such as granting access to a location or software at inference time.
In the enrollment phase, unlike standard face recognition methods, our DNN does not generate and save embeddings in a database but generates a tiny DNN that is sent to the users' edge devices. In more detail, a user enrolls to the system by taking one facial image $\hat{x}^i_{enroll}\in \mathbb{R}^{H\times W\times 3}$, where $H, W$ are the input width and height, and 3 is the color dimension. The image is pre-processed using standard face alignment, as in \cite{kim2022adaface}. The preprocessed image, $x^i_{enroll}\in \mathbb{R}^{h\times w\times 3}$, is sent to the feature extractor $e^i = h_{bb}(x^i_{enroll})$, which is a frozen pre-trained face recognition model that outputs an embedding tensor $e^i\in \mathbb{R}^{d}$, where $d$ is the embedding dimension and depends on the pre-trained model. The generative network $h_{gen}$ is given $e^i$, which represents the user face characteristics as an input and is capable of transforming $e^i$ into personal neural network weights $\theta^i$ that are highly discriminative:
\begin{equation}
    \theta^i = h_{gen}\circ h_{bb}(x^i_{enroll}),
\end{equation}
 where $\theta^i$ helps construct the private model :$f(x,\theta^i)$, and $x$ is some input facial image.

\subsection{Inference Phase}
The inference phase describes the actions of the model when a data stream is fed to it.
Our inference phase is conducted on the user edge device and requires only the computational burden of $f(\theta^i)$.
A facial image is streamed to the model for verification. Then, it is preprocessed in the same manner as in the enrollment phase. The preprocessed image $x_{stream}\in \mathbb{R}^{h\times w\times 3}$ is then utilized using $y = f(x_{stream},\theta^i)$. The output, $y$, is a binary decision on whether $x_{stream}$ is an image of user $i$.
\begin{figure*}[h]
\vspace{0 in}
\begin{center}
\includegraphics[width=.7\textwidth,height=.3\textheight]{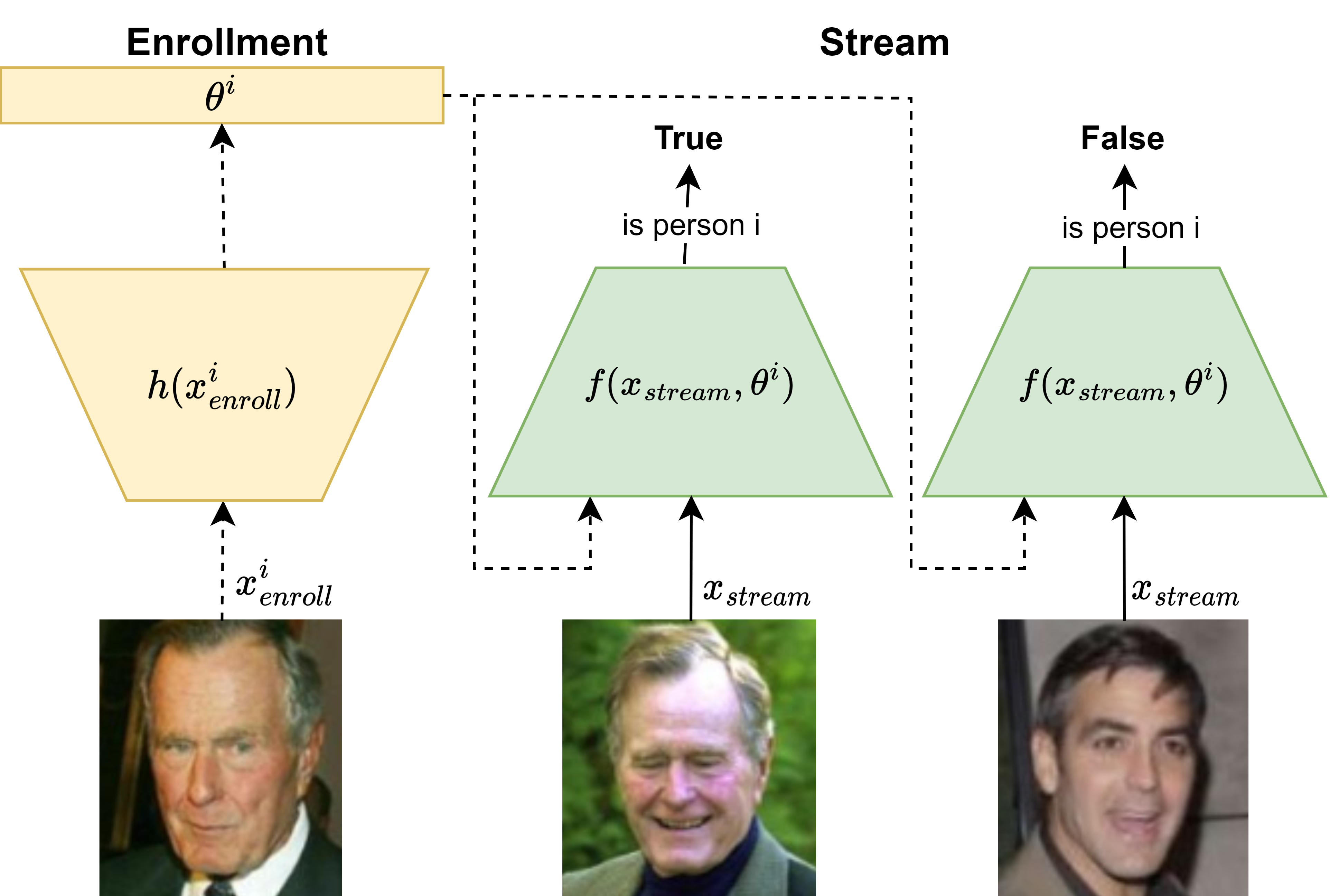}
\end{center}
\caption{Illustration of the inference phase of our method. Enrollment is performed once per user $i$, using a single facial image $x^i_{enroll}$. The enrollment output is a set of weights $\theta^i$ of $f$, designed to fit low-power devices. In the streaming phase, face images $x_{stream}$ are presented to $f$, which then decides whether those belong to person $i$. In the stream phase, $h$ is discarded; thus, the main computational burden is mitigated.}
\label{fig:inferece}
\end{figure*}
\subsection{Training Phase}
\begin{figure*}[h]
\vspace{0 in}
\begin{center}
\includegraphics[width=.7699\linewidth]{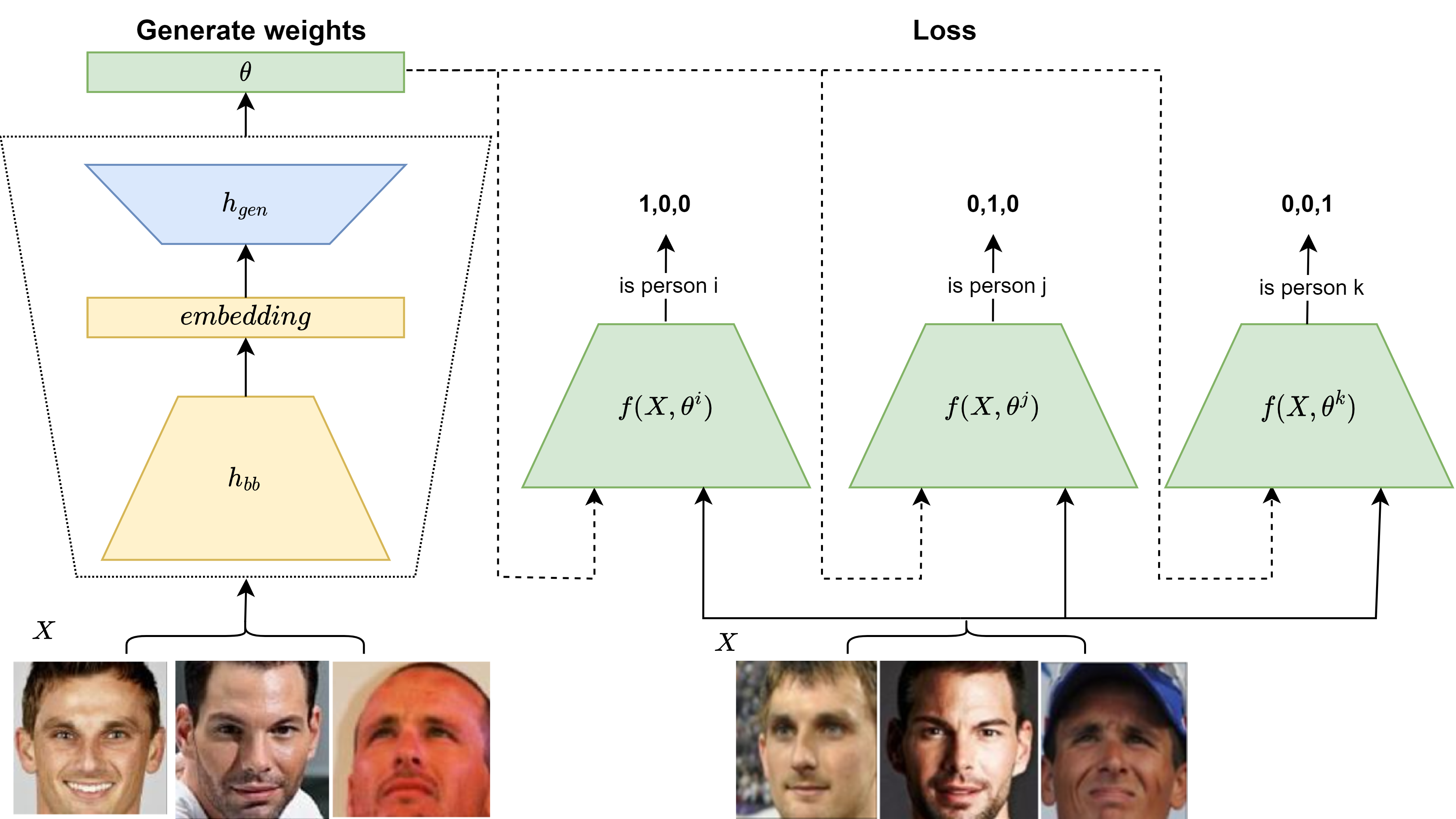}
\end{center}
\caption{The training phase of our method starts with a one-time enrollment procedure, which is noted by dashed arrows. The weights $\theta$ are generated for all images in the batch $X$. Then, $f$ is generated for each user using their specific weights. The streaming phase, noted by regular arrows, may be performed as often as required using the person-specific model $f$. Each model $f$ receives $X$ as an input and calculates the binary cross entropy loss with respect to each model's user label. The embedding model $h_{bb}$ is not trainable. The trainable part $h_{gen}$ obtains the gradients for the batch of $n\times B$ neural networks.}
\label{fig:training}
\end{figure*}

 Training (Fig~\ref{fig:training}) starts by taking an input batch consisting of $nB$ samples from $B$ different identities. Without loss of generality, the following descriptions illustrate the case of balanced sampling, i.e., the number of samples per identity $n$ is constant across the batch. This removes bias towards a specific individual within the batch. Our method also applies to unbalanced sampling, where the constant $n$ can be replaced by a per-user $n^i$. Let us first define an input training batch $X=\{x^1_1,..,x^{1}_{n},x^2_1,..,x^{2}_{n},..,x^{B}_{1},..x^{B}_{n}\}\in \mathbb{R}^{n\times B\times 3\times h\times w}$, where the superscript represents the identity and the subscript denotes the sample index in the batch running from 1 to $n$. 
 At each training iteration, the goal of $h(X)$ is to generate a weights tensor $\Vec{\theta}$, which is $n\times B$ concatenated weight sets corresponding to the input images. Each sample in $j\in X$ results in a weight set $\theta_j$ such that $f(x,\theta_j)$ will recognize only the images of the same identity as the identity of image $j$. 
Let us define an identity mapper that inputs a sample index $j$ and outputs an identity index, formally: $M: j\in[0,nB) \rightarrow i$.
Next, we can define the target matrix $Y\in \mathbb{R}^{nB\times nB}$,
where each row represents the target for sample $j$ as follows 
\begin{equation}
Y_{j,k}=
    \begin{cases}
        1 & M(k)==M(j),\\
        0 & o.w,
    \end{cases}
\end{equation}
with $k\in [0,nB)$. The prediction is constructed in two phases. First, the weight matrix for the batch is created 
\begin{equation}
\Vec{\theta} = h(X).
\end{equation}
$\Vec{\theta}\in \mathbb{R}^{nB \times p}$ where $p$ denotes the number of parameters in $f$. Then, the weights are used to build the personal model for each sample. The second prediction phase can be denoted in a batch form as
\begin{equation}
\hat{Y} = f(X,\Vec{\theta}).
\end{equation}
Where $\hat{Y} \in \mathbb{R}^{nB\times nB}$. We then use the weighted binary cross entropy loss function:
\begin{equation}
\begin{aligned}
L_{bce} = \frac{1}{nB}\sum^{nB-1}_{j=0}\sum^{nB-1}_{k=0}-[\lambda Y_{j,k}log\hat{Y}_{j,k}+\\ (1-\lambda)(1-Y_{j,k}) &log(1-\hat{Y}_{j,k})].
\end{aligned}
\end{equation}

Where $\lambda$ is a factor that compensates for the scarcity of similar identity samples, i.e., since each batch contains $B$ identities, the ratio between positive and negative targets will be $1:B$, $\lambda$ formally defined as:
\begin{equation}
    \lambda = 1-\frac{\beta}{B}.
\end{equation}
For $\beta=1$, we get classic weighted binary cross-entropy, with "perfect" balancing between the classifiers' classes. Nevertheless, $\beta$ should be higher than 1 to reduce future false positives. We used $\beta=2$, but future experiments should attempt even higher values of $\beta$.

Since the $h$ solution space is enormous, i.e., the amount of $\theta$ that $h$ can generate to solve the current batch classification problem, regularizing the output of $h$ is crucial to the generalization of $h$. We propose a weight decay equivalent objective on this set of target models to regularize the weights $\theta^i$ generated by the hypernetwork $h$. The goal is to penalize $h$ when generating large weights, which causes the model to become more noise-sensitive and generalize poorly \cite{jia2022weight}. The objective is defined as:
\begin{equation}
    L_{norm} = \frac{1}{nB} \sum_{j=0}^{n-1}\sum_{0=1}^{B-1}\|h(x_j^i)\|_2^2.
\end{equation}
We evaluate the effect of this regularization in the ablation study performed in Section \ref{sec:experiments}.
Our full objective is a weighted sum of the former losses
\begin{equation}
    L = L_{bce} + \alpha_{norm} L_{norm}\,,
\end{equation}
where $\alpha_{norm}$ is a factor that balances between the two components.

Next, we describe several core components of our training phase.

\noindent{\bf $K$-means centered sampling}
 We now introduce a novel method for effective hard negative mining during batch sampling called $K$-means Centered Sampling (KCS). 
 
\begin{figure*}[t]
\begin{tabular}{cc}
\includegraphics[width=.4575\linewidth,height=4cm]{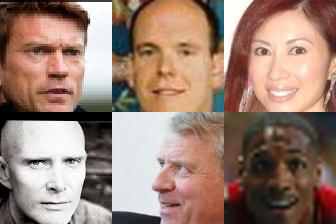} & 
\includegraphics[width=.4575\linewidth,height=4cm]{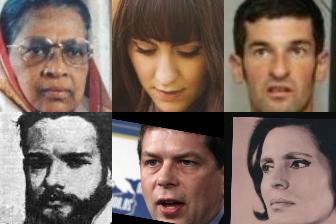}\\
\smallskip
(a)&(b)\\
\smallskip
\smallskip
  \includegraphics[width=.4575\linewidth,height=4cm]{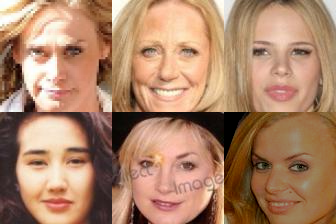}&
 \includegraphics[width=.4575\linewidth,height=4cm]{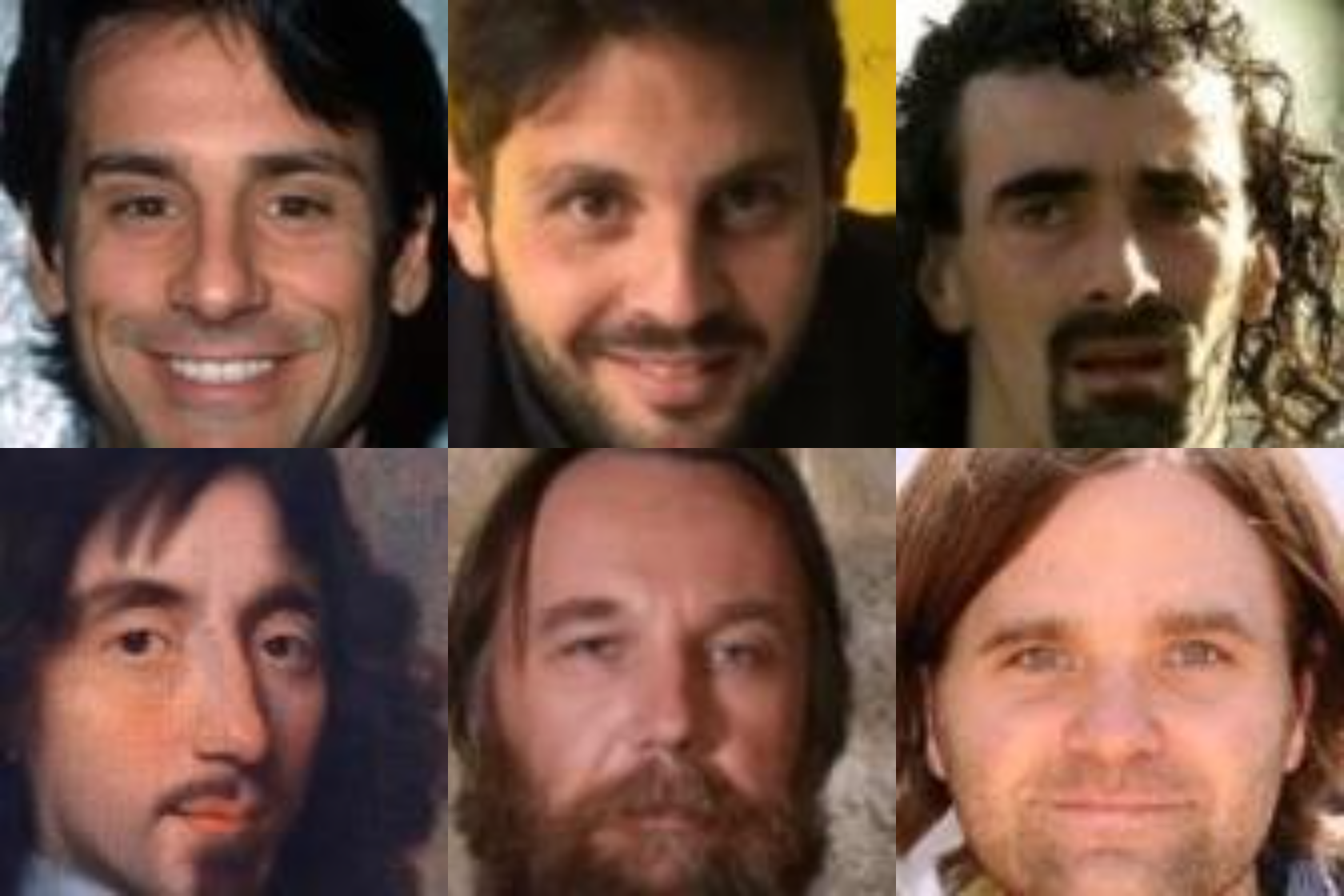}\\
 \smallskip
(c)&(d)\\
\end{tabular}
\caption{Comparing batches obtained with conventional sampling to those obtained with $K$-means centered sampling (KCS). (a, b) a random batch sampled during training, (c, d) a random batch sampled during training when using KCS.}
\label{fig:kcs_qualite}
 \end{figure*}

 In our training, each sample is compared to all other samples in the batch; thus, let us assume uniform sampling over the identities. The identities of most samples in the batch are not necessarily similar and, therefore, may be easy negatives; see Fig.~\ref{fig:kcs_qualite}. In this case, $h$ will quickly minimize $L$ even without learning discriminative features. 
  We want each batch to contain similar identities, forcing $h$ to avoid overfitting and learning robust and discriminative features. To achieve this, we use $h_{bb}$ to extract an embedding for each identity, then cluster these into multiple groups using $K$-means. During training, when employing KCS, batches are sampled from within each cluster.

\noindent{\bf Scheduling}
Training hypernetworks is challenging, especially in fine-grained classification tasks. We utilize scheduling for three hyperparameters to accelerate the model's convergence: batch size, $\alpha_{norm}$, and the epoch in which we start using KCS. Scheduling aims to help the model warm up and make the training problem gradually harder. Since the batch size in our training regime directly affects the difficulty (i.e., the larger the batch, the more complex), we increase the batch size by a factor of 2 several times during warm-up until the final batch size is achieved, which is eight times larger than the initial batch.
Smaller values of $\alpha_{norm}$ lead to fewer constraints on $h_{gen}$ prediction; thus, we start the training with low $\alpha_{norm}$ and increase it gradually.


KCS is designed to construct batches with examples of similar appearance; however, we observed that this often prevents model convergence when used from the initial training phase. To mitigate this, we start by warming up for $T$ iterations using uniform identity sampling and then use KCS during the following iterations.


\noindent{\bf Selection of Architectural Components} The architecture of $h_{gen}$ is simple, comprising elements such as fully connected layers, LayerNorm \cite{ba2016layer}, and GeLU activation \cite{hendrycks2016gaussian}. $f$ predominantly relies on group convolutions for enhancing efficiency and non-parametric normalization layers. Additional information can be found in the implementation details subsection.  

\begin{table*}[htb!]
\caption{Accuracy results on six leading face verification datasets. We present the verification accuracy results, the number of parameters of the inference model, and the number of parameters of the model responsible for the user's enrollment. An identical number of parameters indicates that the enrollment and inference models are identical. We further show the MFLOPS (million floating point operations) for the inference step. }
\centering
\renewcommand{\arraystretch}{1.4}
\large
\adjustbox{max width=1\textwidth}{
\begin{tabular}{lcccccccccc}
\toprule
  Model&Enroll Param.(M) &Infer Param.(M) &MFLOPs &LFW
&CA-LFW
&CP-LFW
&CFP-FP
&AgeDB-30&VGG2-FP \\
 \midrule
  EfficientNet-B0 \cite{tan2019efficientnet} & 5.3&5.3 &-& 99.50 &95.32& 89.73	 &93.47 &95.45&93.20\\
  GhostFaceNetV1-2 \cite{alansari2023ghostfacenets}& 4.06&4.06 &60.29& 99.65 &95.53& 88.57 &91.51 &96.18&-\\
  MobileFaceNetV1 \cite{martinez2021benchmarking} & 3.4&3.4 &1100& 99.40 &94.47& 87.17 &95.80 &96.40&-\\
  ProxylessFaceNAS \cite{martinez2021benchmarking} &3.2&3.2&900&99.20&92.55&84.17&94.70&94.40&-\\
  PocketNetM-256 \cite{boutros2022pocketnet} &1.75&1.75& 1099& 99.58& {\bf 95.63} &90.03 &95.66& {\bf 97.17}&-\\
  PocketNetM-128 \cite{boutros2022pocketnet} &1.68& 1.68& 1099& {\bf 99.60}& 95.67 &90.00 &95.07& 96.78&-\\
   PocketNetS-256 \cite{boutros2022pocketnet} &0.99&0.99&587 & 99.66&  95.50 &88.93 &94.34& 96.35&-\\
  PocketNetS-128 \cite{boutros2022pocketnet} &0.92& 0.92& 587&  99.58& 95.48 &88.63 &94.21& 96.10&-\\
  ConvFaceNeXt 2A \cite{hoo2022convfacenext} & 1.05 & 1.05 & 404.57 & 99.12 & 92.78 & 85.45 & 98.84 & 93.05 &- \\
  ConvFaceNeXt PE \cite{hoo2022convfacenext} & 1.05 & 1.05 & 404.57 & 99.10 & 93.32 & 85.45 & {\bf 98.87} & 92.95 &-\\ 
  ConvFaceNeXt - LCAM \cite{hoo2023lcam} & 1.05 & 1.05 & 406.56 & 99.20 & 93.47 & 86.40 & 89.49 & 93.70 & 90.78  \\
  ConvFaceNeXt - L5K \cite{hoo2023lcam} & 1.05 & 1.05 & 406.60 & 99.23 & 93.73 & 86.97 & 89.61 & 93.65 & 90.04 \\
  ShuffleMixFaceNet-XS \cite{boutros2021mixfacenets} &  1.04&1.04 &161.9&99.53&-&-&91.26& 95.62&-\\%
    DG$_0$ \cite{zhao2023dgfacenet} &  0.42&0.42 &50&99.03&93.67&85.48&87.23& -&88.92\\%
    DG$_1$ \cite{zhao2023dgfacenet} &  0.82&0.82 &150&99.60&95.10&88.90&92.51& -&92.32\\%
  MixFaceNet-XS \cite{boutros2021mixfacenets} &1.04& 1.04& 161.9& {\bf 99.60}& - &- &91.09 &95.85 &-\\
  ShuffleFaceNet 0.5x \cite{martindez2019shufflefacenet} &{\bf 0.5}&0.5&66.9&99.23&-&-&92.59&93.22&-\\
  Ours FaceTransformer\cite{zhong2021face} &145.1& {\bf 0.023}& {\bf 5.4} & 99.52& 94.61 &89.63 &94.96 &93.83&92.96 \\
  Ours AdaFace\cite{kim2022adaface} &145.1& {\bf 0.023} & {\bf 5.4}& 99.55& 94.67 &{\bf 90.58} &95.56 &94.50&{\bf 93.73} \\
\bottomrule
\label{tab:main_results}
\end{tabular}
}
\end{table*}
 
\section{Experiments}
\label{sec:experiments}

\noindent{\bf Implementation details} We use PyTorch \cite{paszke2019pytorch} as a framework. To show that our method is backbone agnostic, we will present results using two backbones (i.e., $h_{bb}$) from different architecture families: Adaface \cite{kim2022adaface} (CNNs) and FaceTransformer\cite{zhong2021face} (Transformer). Adaface \cite{kim2022adaface} uses a ResNet101 \cite{he2016deep} backbone pretrained on the MS-Celeb-1M V2 \cite{deng2019arcface} dataset, and it is frozen for the entire training. We trained $h_{gen}$ using SGD with a learning rate of $1e-3$, momentum value of 0.9, weight decay of $1e-4$, and 3000 warm-up steps. $h_{gen}$ consists of 4 linear layers with LayerNorm \cite{ba2016layer} and GeLU activation \cite{hendrycks2016gaussian}. The variation based on the FaceTransformer \cite{zhong2021face} was also pretrained on the MS-Celeb-1M V2 \cite{deng2019arcface} dataset and was frozen for the entire training. We trained $h_{gen}$ using SGD with a learning rate of $5e-3$ momentum value 0.9, weight decay of $1e-4$, and 3000 warm-up steps. The output of $h_{bb}$ are latent vectors $e$ with size:$B\times T\times 512$, where $T$ is the number of tokens; we therefore first transpose $e$ and then pass it through a series of $1D$ convolutions till achieving the form of $e\in R^{B\times 512}$, which is passed through 4 linear layers with LayerNorm \cite{ba2016layer} and GeLU activation \cite{hendrycks2016gaussian} blocks, which extract $\theta$. We use $K=150$ in $K$-means clustering for KCS. KCS is initiated after 100k steps.
The FLOPs count calculation was conducted using 'thop' \footnote{https://pypi.org/project/thop/} and 'pthflops' \footnote{https://pypi.org/project/pthflops/} Python packages. 

 \noindent{\bf Mini Batch of NNs On a mini-batch of data} 
 Some works that train hypernetworks \cite{shamsian2021personalized} use an inner optimization loop to optimize the parameters $\theta$ for $f$ and use the result as supervision for $h_{gen}$. The downside of this methodology is that, currently, no framework supports backward propagation on a batch of neural networks. This forces a loop over all the batch items and optimization over $\theta_i \forall i$ in the batch. As described in this work, the proposed recipe is different in that it relies on the final classification signal. This is because $f$ is a tiny neural network that reduces the concern of vanishing gradients. Nevertheless, at each forward pass, we need to make a forward and a backward pass-through 
 $nB$ neural networks $f(\theta)$, with each pass containing $nB$ samples. Looping over $nB$ neural networks would make the training procedure almost impractical; thus, we would like to tensorize this procedure. This requires using the \textit{functorch} library \cite{he2021functorch}, which enables forward propagation on a batch of $(nB)^2$ images through a batch of $nB$ neural networks, resulting in a fast and efficient training regime.

 \noindent{\bf Datasets}
Evaluation of our method is done on six widely accepted facial recognition benchmarks. Specifically, we use Labeled Faces in the Wild (LFW) \cite{huang2008labeled} and its later variants CP-LFW \cite{zheng2018cross}, and CA-LFW \cite{zheng2017cross}. We also evaluate our method on CFP-FP \cite{sengupta2016frontal}, AgeDB-30 \cite{moschoglou2017agedb}, and VGG2-FP \cite{cao2018vggface2}.
The MS-Celeb-1M V2 \cite{deng2019arcface} dataset is used to train our models. It is a refined version
of the MS-Celeb-1M \cite{guo2016ms}, containing 5.8M images of 85K identities. 
    
    

 \noindent{\bf Comparison to other methods} 
 We compare our method to 9 leading efficient face recognition models and their multiple variants. All methods were trained using the MS-Celeb-1M V2 \cite{deng2019arcface} training set. Some schemes trained on the WebFace260M dataset \cite{zhu2021webface260m} were dropped for fair comparison \cite{george2023edgeface}. 
 
\begin{table*}[htb!]
\caption{Ablation study demonstrating the importance of different components in our method. Training a model with the same structure as the private model $f$ directly without it being generated by $h$ fails to achieve competitive results. We do not see a benefit in training multiple $h_{gen}$ input layers, leading to worse results than using only the embedding layer. Further, KCS has proved beneficial, improving the results significantly over all datasets. Lastly, using a larger model has not shown improvement over the minimal version.}
\centering
\renewcommand{\arraystretch}{1.4}
\large
\adjustbox{max width=1.\textwidth}{
\begin{tabular}{lcccccccccc}
\toprule
  Model&Enroll Param.(M) &Infer Param.(M) &MFLOPs &LFW
&CA-LFW
&CP-LFW
&CFP-FP
&AgeDB-30&VGG2-FP & Avg. \\
 \midrule
 Training $f$ directly  & 0.023 & 0.023 & 5.4 & 90.70 & 75.17& 65.72& 68.21 & 72.70 &71.60  & 74.02 \\
 Multiple $h_{gen}$ input layers & 147.2 & 0.023 & 5.4  &99.41   & 93.48 &88.52  & \textbf{95.59}  & 93.90  &  93.64 &  94.09 \\
Ours AdaFace\cite{kim2022adaface} w/o KCS & 147.2 & 0.023 & 5.4  &99.23   & 93.18 &88.77  & 93.51  & 91.20  & 92.54  & 93.07  \\
Ours AdaFace\cite{kim2022adaface} w/o $L_{norm}$ & 147.2 & 0.023 & 5.4  &99.29   & 93.37 &89.11  & 93.61  & 91.79  & 92.82  &  93.33 \\
 Ours AdaFace\cite{kim2022adaface} &118.6& 0.023& 5.4& \textbf{99.55}& \textbf{94.67} &\textbf{90.58} &95.56 &94.50&93.73&\textbf{94.77} \\
 Ours AdaFace\cite{kim2022adaface}, larger model &147.4& 0.037& 5.7 & 99.50& 94.60 &90.45 &94.80 &\textbf{94.77}&\textbf{93.84}&94.66 \\
 \bottomrule
 \label{tab:ablation}
\end{tabular}
}
\end{table*}

\subsection{Results}
\begin{figure*}[t]
 \vspace{0.15 in}
\begin{tabular}{cc}
 \includegraphics[width=.48\linewidth]{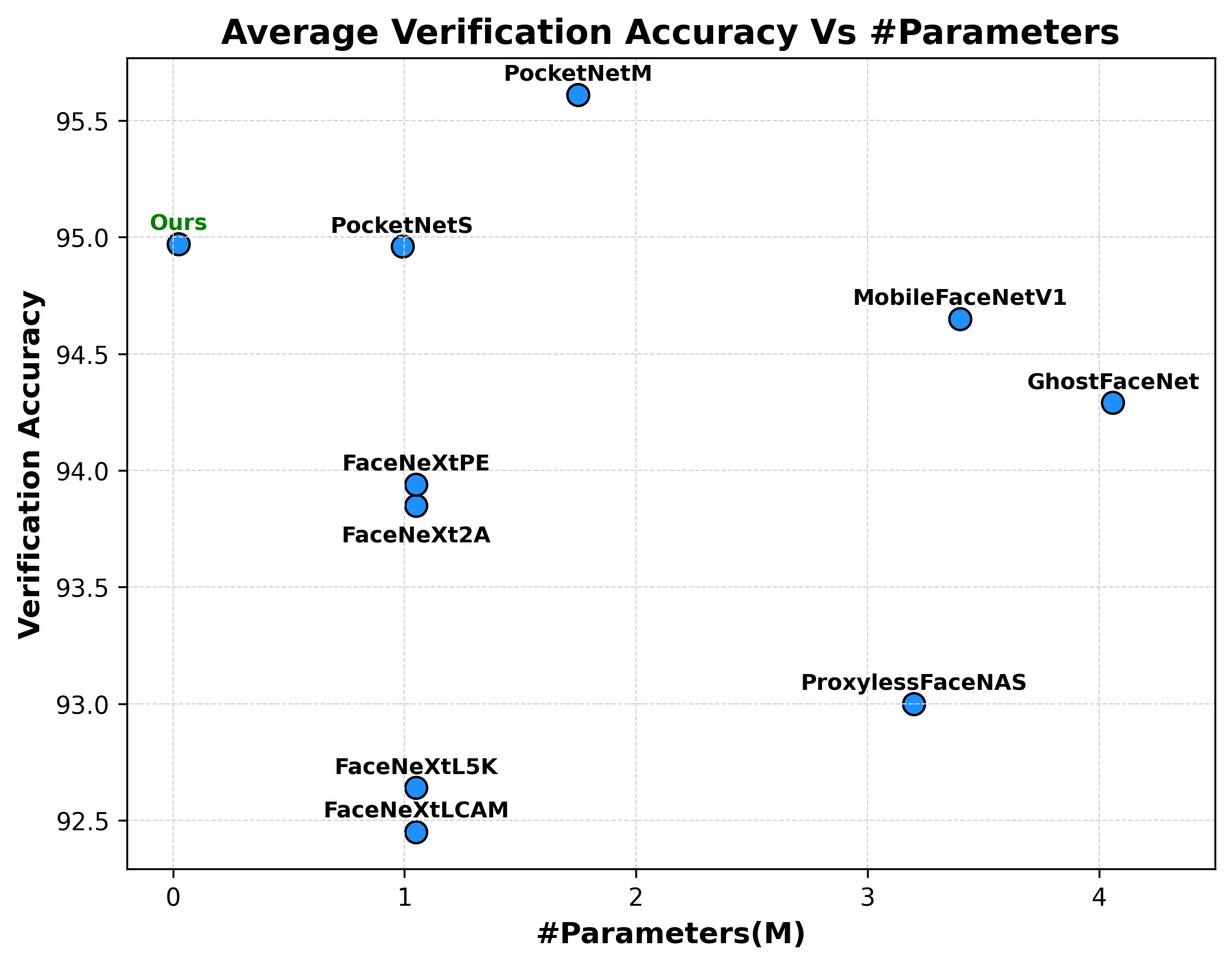}&
 \includegraphics[width=.48\linewidth]{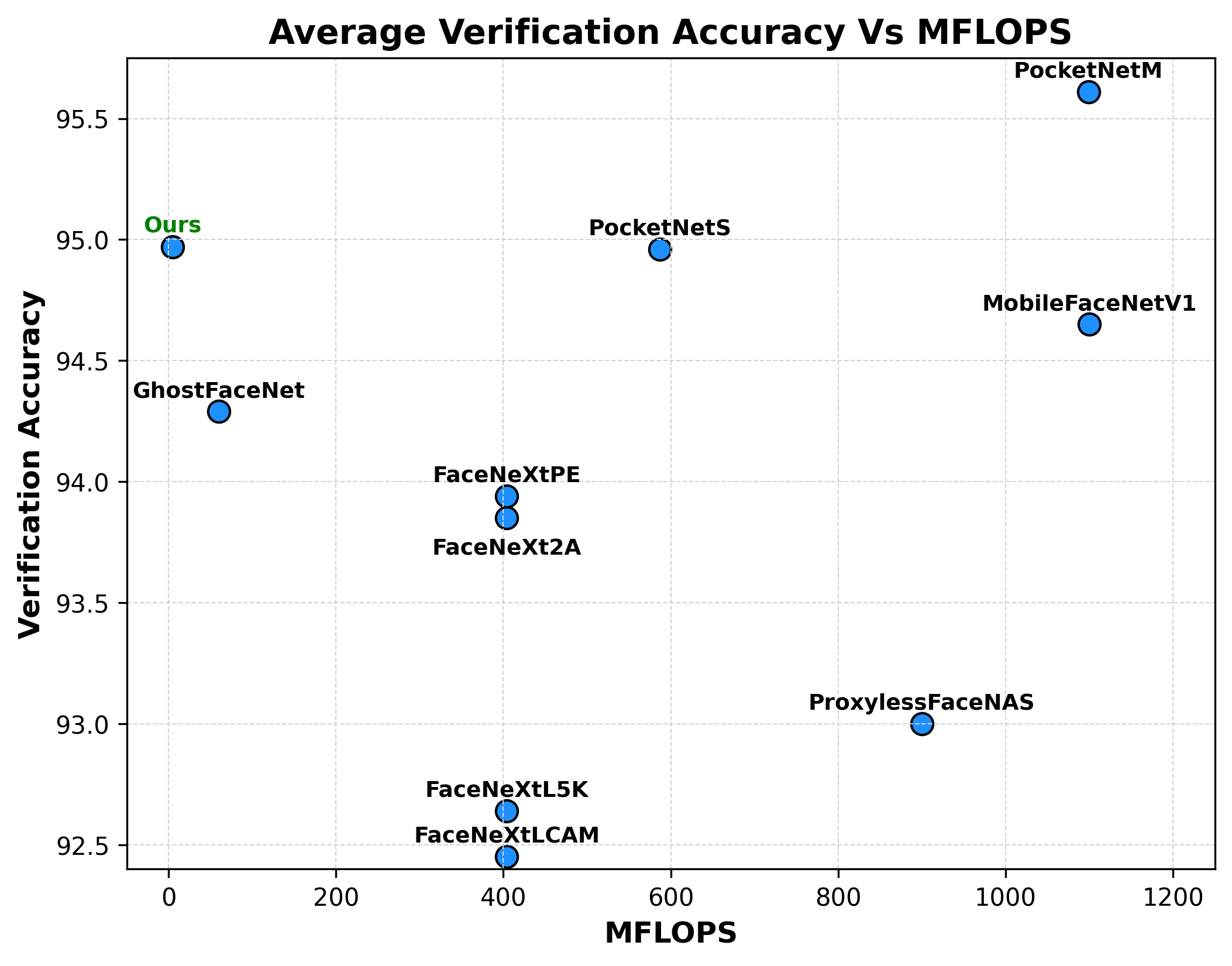}\\
(a) & (b)\\
\end{tabular}
\caption{(a) The top1 accuracy average vs. the number of parameters for five datasets (excluding vgg2-fp) of our models and efficient competing models for face recognition. (b) the same when comparing accuracy and the number of operations in mega-flops.} 
 \vspace{0 in}
\label{fig:flops_params_acc}
\end{figure*}

Table \ref{tab:main_results} presents verification accuracy results on the six leading face verification datasets mentioned above. Although our method requires a large enrollment model, this is used only once per user, and only the inference model is deployed on the edge device. When comparing inference model size, i.e., its number of parameters and amount of MFLOPs required, we can see the significant advantage of our model. Although our model is almost 22 times smaller than its smallest competitor \cite{martindez2019shufflefacenet}, it maintains on par or better results on all available datasets when the two are compared. Even compared to more than 100 times larger models, we still obtain state-of-the-art results on the CFP-FP dataset. For the first time, we demonstrate a method that dramatically reduces memory requirements while maintaining reasonable results on multiple benchmarks. Our method is based on a state-of-the-art face recognition backbone for the initial embedding during enrollment. It is improved when using the more accurate backbone (AdaFace \cite{kim2022adaface} instead of FaceTransformer\cite{zhong2021face}). Still, the improvement is slight, and both backbones are comparable. A visual overview of the results on the five datasets: LFW, CA-LFW, CP-LFW, CFP-FP,
and AgeDB-30, is shown in Fig.~\ref{fig:flops_params_acc}, only the methods that have results on all five datasets are presented. Fig.~\ref{fig:flops_params_acc} shows both the verification accuracy vs. the number of parameters and the number of floating point operations in mega-flops. We expect further improvements when using backbones trained on cleaner and larger datasets, such as subsets of the WebFace260M dataset \cite{zhu2021webface260m}, similarly to the improvement obtained by \citet{kim2022adaface} when training on larger datasets. 

\noindent{\bf Ablation study} 
 Key parts of our method are examined via an ablation study shown in Table~\ref{tab:ablation}. We first tested the option of training $f$ in a direct manner using the AdaFace \cite{kim2022adaface} training scheme. This resulted in a massive degradation in accuracy, which can be explained by the model underfitting the dataset due to its small number of parameters. Additional ablation was done using multiple layers as input for $h_{gen}$. We have not seen any benefit of using additional feature maps from different layers as input compared to our main result, which uses only a single embedding layer. This may be because more input feature maps for weight creation resulted in a larger $h_{gen}$, which overfitted the data and worsened performance. 
 
 Another experiment was done using a model $f$ with a larger number of parameters, which led to on-par results. We assume that the limiting factor to our model's performance is not the number of parameters but the procedure of generating those; we will discuss the limiting factors of our method and how to solve them in the Discussion section \ref{sec:discussion}
 
 Lastly, we check the impact of removing our KCS training technique and the $L_{norm}$ loss term. In both cases, this led to a significant decrease in accuracy; the largest decrease was on the AgeDB-30 dataset, which is the hardest benchmark. 

\section{Discussion}
\label{sec:discussion}
Our method suffers from a few limitations. First, it converges quickly to a decent performance, but obtaining competitive results could be time-consuming. 
We suspect that our cosine learning rate scheduler is not optimal for this optimization task. 

Second, from the efficiency viewpoint, increasing the batch size results in a quadratic increase in training complexity. This happens because each sample is compared to all other samples in the batch, which is computationally intensive. Exploring a way of training with a smaller batch size for a longer time without hurting the convergence quality may prove beneficial. Alternatively, one can select a subset of the pairs in the batch. 

Third, previous works in generative models showed that methods that generate with iterative updates \cite{dhariwal2021diffusion} are better than one-step generation models, such as GANs \cite{goodfellow2014generative}. 
 This suggests that generating the $\theta^i$ in a few steps, for example, using a diffusion process, \cite{lutati2023ocd} may be beneficial. 

Currently, $f$ has some drawbacks; it could be improved using conventional normalization methods, such as batch or layer normalization. Future work could include training on larger datasets, such as WebFace4M/12M \cite{zhu2021webface260m} or larger. This will aid us in leveraging larger and stronger $h_{gen}$ architectures. 

While our current work focuses on creating a model from a single enrollment image per user, we could explore using more images if they are available. In cases where multiple images are available, we can generate multiple enrollment images, embedding them and then averaging them as an input to $h_{gen}$. Such an approach can increase the robustness of $f$ to different views, lighting, and more. 

Further improvements may consider data augmentations as a tool to enhance the input data. For example, using geometric augmentations such as flip, crop, rotate, or translate the face images to augment the single image:$x_{enroll}^i$ and average the embedding to make the input of $h_{gen}$ more robust. 
\section{Conclusions}
\label{sec:conclusions}
This work presented an efficient paradigm for running efficient DNN on edge, not by suggesting a more efficient model but by changing the problem scope being solved. The change we suggested is treating the cases of very small databases, such as smartphones, tablets, and personal laptops, differently from large databases. 
At the core of our work stands the idea of a simpler problem with a more straightforward solution. Our solution no longer needs a very general neural network but one that solves a particular problem. This significantly reduced the number of parameters while achieving on-par accuracy for multiple datasets. Although this paper focuses on face verification, we believe our method can also be applied in other fields, such as speaker identification and keyword spotting. 

\clearpage
{\small
\bibliographystyle{IEEEtranN}
\bibliography{egbib}
}

\end{document}